\documentclass[letterpaper, 10 pt, conference]{ieeeconf}  
\usepackage{amsmath,amsfonts,bm}
\usepackage{mathtools}
\usepackage[bb=boondox,bbscaled=.95]{mathalfa}
\usepackage{cancel}
\usepackage{changes}
\usepackage{hyperref}

\usepackage{tikz}
\usepackage{pgfplots}
\usepgfplotslibrary{groupplots}
\usepgfplotslibrary{patchplots}
\pgfplotsset{compat=1.14}

\usepackage{bm}

\DeclareMathOperator{\diag}{diag}

\DeclareMathOperator{\Nat}{\mathbb{N}}

\DeclareMathOperator{\C}{\mathcal{C}}

\DeclareMathOperator{\Lss}{\mathcal{L}}

\DeclareMathOperator{\Sa}{\mathcal{S}}
\DeclareMathOperator{\T}{\mathcal{T}}

\DeclareMathOperator{\vep}{\varepsilon}
\DeclareMathOperator{\vz}{\mathbb{0}}
\DeclareMathOperator{\mz}{\mathbb{O}}
\DeclareMathOperator{\mi}{\mathbb{I}}

\newcommand{\y}{\mathbf{y}}

\tikzset{
	NNnode/.pic={
		\pgfmathsetmacro\RecH{2}
		\pgfmathsetmacro\RecW{\RecH/10}
		\coordinate (-ll) at (-\RecW/2,-\RecH/2);
		\coordinate (-ur) at (\RecW/2,\RecH/2);
		\coordinate (-lr) at (-ll-|-ur);
		\coordinate (-ul) at (-ll|--ur);
		\path (-ul) -- (-ur) coordinate[midway] (-north);
		\path (-ll) -- (-lr) coordinate[midway] (-south);
		\path (-ll) -- (-ul) coordinate[midway] (-west);
		\path (-ur) -- (-lr) coordinate[midway] (-east);
		
		\begin{scope}[shift={(-\RecW/2,-\RecH/2)}]
			\draw (-ll) rectangle (-ur);
			\foreach \y in {0.05,0.5,0.75,0.85,0.95}
			\draw (0.5*\RecW,\RecH*\y) circle[radius=0.3*\RecW];
			\foreach \y in {0.275,0.625} {
				\fill (\RecW*0.4,\y*\RecH-0.1*\RecW) rectangle (0.6*\RecW,\y*\RecH-0.3*\RecW);
				\fill (\RecW*0.4,\y*\RecH+0.1*\RecW) rectangle (0.6*\RecW,\y*\RecH+0.3*\RecW);
			}
		\end{scope}
	}
}

\newtheorem{proposition}{Proposition}
\newtheorem{remark}{Remark}

\newtheorem{definition}{Definition}

\newcounter{assc}
\allowdisplaybreaks
\DeclareMathOperator{\R}{\mathbb{R}}

\newcommand{\cM}{\mathcal M}

\newcommand{\cU}{\mathcal U}

\newcommand{\cW}{\mathcal W}

\newcommand{\x}{\times}

\newcommand{\dw}{{n_w}}

\IEEEoverridecommandlockouts                              
\title{\LARGE \bf
    Stable Neural Flows
}

\author{Stefano Massaroli$^{1,\star}$, Michael Poli$^{2,\star}$, Michelangelo Bin$^{3,\star}$, \\
Jinkyoo Park$^{2}$, Atsushi Yamashita$^{1}$ and Hajime Asama$^{1}$
\thanks{$^{1}$Stefano Massaroli is with the Department of Precision Engineering, the University of Tokyo, Tokyo, Japan
        {\tt\small massaroli@robot.t.u-tokyo.ac.jp}}%
\thanks{$^{2}$Michael Poli is with the Department of Industrial Engineering, Korean Advanced Institute of Science and Technology, Daejeon, South Korea
        {\tt\small poli.m@kaist.ac.kr}}
\thanks{$^{2}$Michelangelo Bin is with the Department of Department of Electrical and Electronic Engineering, Imperial College London, London, United Kingdom}
\thanks{$^{\star}$Equal contribution authors}
}
\begin{document}
\maketitle
\thispagestyle{empty}
\pagestyle{empty}

\begin{abstract}

We introduce a provably stable variant of \textit{neural ordinary differential equations} (neural ODEs) whose trajectories evolve on an energy functional parametrised by a neural network. \textit{Stable neural flows} provide an implicit guarantee on asymptotic stability of the depth--flows, leading to robustness against input perturbations and low computational burden for the numerical solver. The learning procedure is cast as an optimal control problem, and an approximate solution is proposed based on adjoint sensivity analysis. We further introduce novel regularizers designed to ease the optimization process and speed up convergence. The proposed model class is evaluated on non--linear classification and function approximation tasks.
\end{abstract}
\section{INTRODUCTION}
Neural networks are function compositions of the form
\[
    u \mapsto f_S\circ\cdots\circ f_0(u) := \phi_S(u)
 \]
which realize a possibly very complex nonlinear mappings between input and output spaces. Recent works \cite{sonoda2017double,lu2017beyond,chen2018neural,sonoda2019transport}, explored the continuum limit of neural networks where the input--output mapping is realised by the solution (flow) of an ordinary differential equation $\frac{dx}{ds} = f(s, x(s)),~x(0) = u$, defined on a compact \textit{depth domain} $\Sa\subset\R$, where $s\in\Sa$ denotes the \textit{depth} variable. At its core, this approach turns the task of \textit{learning} an input--output map into a data--driven search for a suitable vector field $f$. In this context, $f$ is often parametrised by a neural network $f:=f(s, x(s),w(s))$ in which $w\in\mathcal{W}$. \textit{Neural ordinary differential equations} (neural ODEs) have been successfully used as building blocks for more elaborate data--driven models \cite{rubanova2019latent,yildiz2019ode2vae} and the formulation has been adapted to support other classes of differential equations, e.g. hybrid systems \cite{jia2019neural} or \textit{stochastic differential equations} (SDE) \cite{li2020scalable}. While \cite{chen2018neural} models $f$ as multi--layer feed--forward networks and discrete convolution operators, the framework has also been extended to spectral graph convolutions and networked models \cite{poli2019graph}.
Neural ODEs have also been shown to improve upon previous learning methods in the approximation of physical systems \cite{greydanus2019hamiltonian,rackauckas2020universal}. 

Although the framework has seen application in machine learning tasks, the lack of a rigorous derivation of the optimization process and stability guarantees prevent its widespread use in control applications. The unconstrained form of $f$ may yield unstable and \textit{stiff} dynamics, which in turn leads to an increase in sensitivity to input perturbations \cite{massaroli2020dissecting} and computational overheads of numerical solvers. Furthermore, as highlighted in \cite{massaroli2020dissecting}, unconstrained neural ODEs can give rise to chaotic behaviors. The sensitivity to small deviations of the input data features (e.g. \textit{adversarial attacks}) renders these models dangerous in practice: although they may fit training data, generalization to unseen data becomes unreliable due to error propagation.

Imposing stability in discrete neural networks as a first--order design principle has lead to a variety of high performance model variants \cite{haber2017stable, chang2019antisymmetricrnn}. Within the neural ODE framework, heuristic approaches to stability \cite{hanshu2019robustness} have been experimentally shown to improve robustness of neural ODEs. However, these approaches do not provide stability guarantees and require tuning a specific regularization term. In this work, we introduce a \textit{provably} stable neural ODE variant, \textit{stable neural flows}, whose trajectories evolve on monotonically non--increasing level sets of an energy functional parametrised by a neural network. The training task is approached as an optimal control problem, and a recursive adaptation procedure is specifically developed to approximate its solution. The proposed adaptation law is based on a gradient descent procedure covering both temrinal and backpropagated settings. Finally, we introduce \textit{ad hoc} regularizers to ease the optimization process for the proposed model.

\begin{figure}
    \centering
    \includegraphics{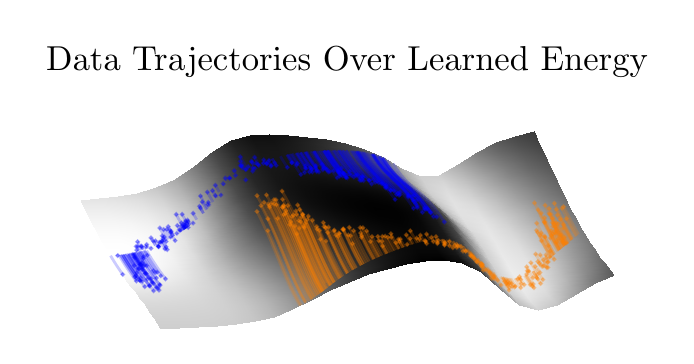}
    \vspace{-5mm}
    \caption{Trajectories of the \textit{half moons} data over learned energy functional, dissipated along the flows.}
    \vspace{-5mm}
    \label{fig:my_label}
\end{figure}
\subsubsection*{Notation}
$\Nat$ and $\R$ ($\R^+$) are the sets of natural and real (positive real) numbers; $\C^n$ is the set of n-times continuously differentiable functions. Moreover, $\dot x:=\frac{dx}{ds}$ and $\partial$ denotes the transposed gradient operator $\partial_x:={\partial^\top}/{\partial x}$. $\mathbb{E}[\cdot]$ is the \textit{expected value} operator. $\vz_n$ is the origin of $\R^n$, $\mz_{n\x m},~\mi_{n}$ are the $n$--by--$m$ null matrix and $n$th order and identity matrix, respectively.
\section{Stable Neural Flows}
In this section we introduce \textit{stable neural flows} as a stable variant of neural ODEs. In particular, we define a function $\vep$, which we call \textit{energy function}. $\vep$ is parametrised by a neural network and steers the flows of the state $x$ be steered along its negative gradient directions. First, we start with some necessary preliminary concepts, which will be followed by the definition of our proposed model class.
\subsection{Preliminaries}
Let $(\T,\geq)$ be a linearly ordered set, called the \textit{data time} set (typically $\T=\Nat$). We suppose to be given an \emph{input-output data stream}, which is a net of the form $\big\{ (u_t,y_t)  \big\}_{t\in\T}$ of input-output pairs $(u_t,y_t)\in\R^{n_u}\times\R^{n_y}$. 

\begin{definition}[Neural ODE]
With $h_u:\R^{n_u}\rightarrow\R^{n_x}$, $h_y:\R^{n_x}\rightarrow\R^{n_y}$ two affine maps, called the \textit{input projection} and \textit{output projection} respectively, a neural ODE is a system of the form
\begin{equation}\label{eq:node}
    \begin{aligned}
        &\dot x =  f\left(u_t, x(s), w(s)\right)\\
        & x(0) = h_u(u_t)\\
        & \hat y_t = h_y(x(S))\\
    \end{aligned}~~~s \in \Sa
\end{equation}
where $\Sa:=[0,S]$ ($S\in\R^+$) is the \textit{depth domain} and $f:\R^{n_u}\x\R^{n_x}\x\mathcal{W}\rightarrow\R^{n_x}$ is a neural network with weights $w\in\mathcal{W}$, being $\cW$ a given pre-specified class of functions $\Sa\to\R^\dw$.
\end{definition}

At each $t\in\T$, system \eqref{eq:node} takes as input  $u_t$ and produces as output its \emph{best  estimate} $\hat y_t$ of the corresponding output  $y_t$.  Our degree of freedom in  \eqref{eq:node} is the choice of the parameter $w$ inside $\cW$. Therefore, our model set is the family of systems of the form \eqref{eq:node} obtained as $w$ ranges in $\cW$. Note that, in many practical cases, it is convenient to extend the search of the model by ``learning'' the input and output projections $h_u$ and $h_y$. This results particularly useful when high--dimensional embeddings of the input $u_t$ are contextually required by the specific problem \cite{dupont2019augmented,massaroli2020dissecting}. 

Within the scope of this paper we only limit our analysis to \textit{depth--invariant} weights, i.e. we assume that W is the set of constant functions. For ease of notation, we thus identify $\cW$ with $\R^{n_w}$, and in the following we do not distinguigh between the two.
\begin{remark}[Well--Posedness]\label{rem:wellpos}
    If $f$ is Lipschitz, for each $u_t$ the initial value problem in \eqref{eq:node} admits a unique solution $x$ defined in the whole $\Sa$. If this is the case, there is a mapping $\phi$ from $\R^{n_w}\times\R^{n_u}$ to the space of absolutely continuous functions $\Sa\mapsto\R^{n_x}$ such that
    $x_t:= \phi(w,u_t)$ satisfies the ODE in \eqref{eq:node}. This in turn implies that, for all $t\in\T$, the map
    \begin{equation*}
    (w,u_t)\mapsto \gamma(w,u_t) := h_y\big(\phi(w,u_t)(S)\big)
    \end{equation*}
    satisfies $\hat y_t = \gamma(w,u_t)$.
    For the sake of compactness, we denote $\phi(w,u_t)(s)$ by $\phi_s(w,u_t)$, for any $s\in\Sa$.
\end{remark}
{\color{red}
    
}
\subsection{Stable Neural Flows}
\begin{definition}[Stable neural flow]
    A \textit{stable neural flow} is a variant of (\ref{eq:node}) having the form
    \begin{equation}\label{eq:snode}
        \dot x = -{\partial_x}\vep(u_t, x(s), w)~~s\in\Sa,\forall t\in\T
    \end{equation}
where $\varepsilon:\R^{n_x}\times\R^{n_u}\times\R^{n_w}\rightarrow\R$ is a neural network which is $\C^\infty$ and bounded from below.
\end{definition}

Under the above assumptions, the solutions of $(\ref{eq:snode})$ are forward complete for any compact depth domain $\Sa$ and $t\in\T$, according to Remark \ref{rem:wellpos}. Moreover, the explicit dependence of the energy function on the input data $u_t$ allows the model to learn a family of energy functionals \textit{ad hoc} for each data point, rather that a single one. This greatly increases the \textit{expressivity} of the model while reducing the stiffness of the differential equation, allowing it to learn complex nonlinear mappings without introducing additional structure \cite{massaroli2020dissecting}.

Stability of the proposed model is assessed by the following:
\begin{proposition}[Stability]\label{prop:stab}
    Every closed set $\cM\in\R^{n_x}$ such that
    \[
        \forall x\in\cM,~\partial_x\vep=\mathbb{0}_{n_x}~\text{and}~\partial_x^2\vep\succ 0
    \]
     which is contained in an open neighborhood $\cU\supset\cM$ satisfying
    \[
        \forall x\in\cU\setminus\cM,~\partial_x\vep\neq\mathbb{0}_{n_x}
    \]
    is locally asymptotically stable.
\end{proposition}
\proof{
    For all $x\in\cM$, $\dot x = \mathbb{0}_{n_x}$ and, thus, $\cM$ is forward invariant. Let $\cU\supset\cM$ be an open neighborhood of $\cM$ such that $\forall x\in\cU\setminus\cM,~\partial_x\vep\neq\mathbb{0}_{n_x}$ and let $V(x) = \vep(x)$ be a candidate Lyapunov function. It holds:
        $\forall x\in\cU\quad\dot V = -\langle\partial_x V,\partial_x V\rangle\leq 0$ and  $\forall x\in\cM~~\dot V = 0$.
    %
    Thus $\cM$ is asymptotically stable.
    }
\endproof
\begin{remark}
    From Prop. \ref{prop:stab}, it follows that every isolated set $\cM$ of local minima of $\vep$, has a basin of attraction. Thus, for almost all initial conditions, the system will dissipate all the energy reaching some stable set. In fact, for all initial conditions $h_u(u_t)\in\R^{n_x}$ which are not local maxima of $\vep$ nor belong to the basins of attraction of its saddle points, there exists a stable set $\cM$ such that $\lim_{s\rightarrow\infty}\phi_s(w, u_t)\in\cM$.
\end{remark}
{

Note that, even in a compact $\Sa=[0,S]$ we can seek a suitable set of parameters $w$ to reach an arbitrary level of steady state (i.e. closeness to some invariant $\cM$).}
\subsection{Variants of the Model}
\subsubsection{Port--Hamiltonian inspired model}
    By noticing that (\ref{eq:snode}) is formally an autonomous \textit{port--Hamiltonian} system \cite{ortega2001putting, massaroli2019port}, we can generalise the model to 
    \begin{equation}\label{eq:ph}
        \begin{aligned}
        \dot x = A(x,w_{A})\partial_x\varepsilon(u_t,x,w_{\vep}),\quad w:=(w_A,w_{\vep})
        \end{aligned}
    \end{equation}
    where it is easy to prove that stability, i.e. dissipativity of $\varepsilon$, is preserved whenever $A$ is chosen such that $A(x)+A^\top(x)\prec 0$ for any $x$. Note that such an $A$ gives more ``freedom'' to the model by combining different (negative) gradient directions to effectively steer the states towards some energy minima.
\subsubsection{Second--order model}
    A special case of stable neural flows modeling is to mimic the classical Hamiltonian dynamics of a mechanical system as follows. Let $x: = (q,p)\in\R^{n_x}$, $p,q\in\R^{n_v}$ ($n_v=n_x/2$). A second--order stable neural flows model can be defined as
    \begin{equation}\label{eq:s2node}
        \begin{aligned}
            & \dot q = p \\
            & \dot p = -\alpha p - \partial_q\varepsilon(u_t, q(s), w)
        \end{aligned}
    \end{equation}
    where $\alpha\in\R^+$ is a trainable parameter. Within this framework, stability can be proven by defining a total energy function $\varphi(q,p) = \frac{1}{2}p^\top p + \varepsilon(q, w)$. In fact, it holds
    \begin{equation}
        \begin{aligned}
            \frac{d}{ds}\varphi(q,p) &= 
                \begin{bmatrix}
                    \partial^\top_q\varphi & \partial^\top_p\varphi
                \end{bmatrix}
                \begin{bmatrix}
                    \dot q \\
                    \dot p
                \end{bmatrix}\\
            &=  \partial_q^\top\varepsilon p - \alpha p^\top p - p^\top\partial_q\varepsilon\\
            &= -\alpha p^\top p\leq 0 ~~\forall s\in\Sa
        \end{aligned}
    \end{equation}
    and $(q,p)$ eventually converges to some fixed point $(q^*, \mathbb{0}_{n_v})$ where $q^*$ is a local minimizer of $\varepsilon$. Also in this case, inspired by port--Hamiltonian models, we can define a stability--preserving generalization of (\ref{eq:s2node}) as
    \[
        \begin{bmatrix}
            \dot q\\
            \dot p
        \end{bmatrix} 
        = 
        \begin{bmatrix}
            \mathbb{O} & B\\
            -B & A
        \end{bmatrix}
        \begin{bmatrix}
            \partial_x\varepsilon\\
            p
        \end{bmatrix} 
    \]
    where $A=A^\top\preceq 0$ and $B=B^\top$.
\subsubsection{Stochastic Model}
  Motivated by the stochastic variants of neural ODEs \cite{jia2019neural} and their growing interest within the research community, we define the \textit{stable stochastic neural flows} resembling a drift--diffusion model
    \begin{equation}
        \begin{aligned}
            & dx = -\partial_x\varepsilon(x(s),w)ds + \sqrt{2\beta^{-1}}dW(s)
        \end{aligned}
    \end{equation}
    where $W(s)$ is a Weiner process. In this context, the energy $\varepsilon(x, w)$ assumes the role of \textit{drift potential} while the diffusion coefficient $\beta\in\R^+$ is a (optimisable) model parameter.
    
    Here, stability can be assessed by recalling the flows of the \textit{joint probability density} $\rho(x,s)$. In fact, $\rho(x, s)$ evolves according to the \textit{Kolmogorov forward} partial differential equation \cite{caluya2019gradient},
    \begin{equation}\label{eq:jpd_pde}
        \partial_s \rho= \partial_x\cdot\rho\partial_x\varepsilon + \beta^{-1}\partial_x\cdot\partial_x\rho
    \end{equation}
    where ``$\cdot$'' denotes the inner product. (\ref{eq:jpd_pde}) have a unique stationary solution $\rho_{\tt ss}(x)$ corresponding to the Boltzmann distribution $\rho_{\tt ss}(x) = \lim_{s\rightarrow\infty}\rho(x, s) = \kappa e^{-\beta\varepsilon(x, w)},~(\kappa\in\R)$ which is reached while the Lyapunov functional $\mathbb{E}_\rho[\varepsilon + \beta^{-1}\log\rho]$ (often called \textit{free energy}) decays along the flow.
    Besides, optimising the parameters of such models requires extending the \textit{adjoint sensitivity analysis} to a stochastic setting, treated e.g. in \cite{elliott1989adjoint} and more recently in \cite{li2020scalable}. The experimental evaluation of this model is therefore out of the scope of the paper and it will be treated in future work.
\vspace{-5mm}
\section{Training The Model}
\vspace{-2mm}
After fixing the model structure, 
our remaining degree of freedom lies in the choice of the parameters $w$. In this work, we cast this task in an optimal control setting. In particular, we define a smooth scalar cost function $\ell(w,x,u_t,y_t)$ measuring how well the model fits the data and, consequently, we optimise the parameters to minimise $\ell$ for all $t\in\T$. This procedure is also referred to as \textit{training} in machine learning terminology. 
From now on, we denote $\ell(w,x,u_t,y_t)$ as $\ell_t$.
\subsection{Training Process: an Optimal Control Perspective}
We start by defining the training for standard neural ODEs. In the following, we assume $\T$ to be a finite set.
The training process can be then defined as the following constrained nonlinear program
\begin{equation}\label{eq:opt}
    \begin{aligned}
        \min_{w\in\R^{n_w}}  &\frac{1}{|\T|}\sum_{t\in\T}\ell_t \\
        \text{subject to}~~
        &\dot x =  f\left(u_t, x(s), w\right)~~s \in [0,S]\\
        & x(0) = h_u(u_t)\\
        & \hat y_t = h_y(x(S))\\
        & \forall t\in\T
    \end{aligned}
\end{equation}
While, in general, it is not possible to obtain an analytic solution to (\ref{eq:opt}), as customary in the pertinent literature, we can approximate a locally optimal value of the model's parameters recursively by \textit{gradient descent} (GD) \cite{sun2019survey}.

Let $k\in\Nat$ be the counter the GD iterations. In case all the input--output data are available offline, the GD solution is obtained by iterating
\begin{equation}\label{eq:gd}
    \begin{aligned}
    w_{k+1} = w_k - \eta\frac{1}{|\T|}\sum_{t\in\T}\frac{d\ell_t}{dw}~~(\eta\in\R^+)
    \end{aligned}
\end{equation}
since $\frac{d}{dw}\sum_{t\in\T}\ell_t=\sum_{t\in\T}\frac{d\ell_t}{dw}$. On the other hand, if the input--output data becomes available sequentially, we have $k=t$ and we implement the \textit{stochastic} version of GD
\[
    w_{t+1} = w_t - \eta\frac{d\ell_t}{dw}
\]
where $t$ is reset to $0$ whenever $t=|\T|$. With a sufficiently small value of $\eta$, and a sufficiently large number of steps, (stochastic) GD converges arbitrary close to a local minimizer of $\frac{1}{|\T|}\sum_{t\in\T}\frac{d\ell_t}{dw}$ \cite{sun2019survey}.

Regardless of the choice of $\ell_t$, the gradients with respect to the ODE parameters $w$ have to be computed. According to \cite{chen2018neural}, gradients can be computed with $\mathcal{O}(1)$ memory efficiency through adjoint sensitivity analysis \cite{pontryagin1962mathematical,cao2002adjoint}. In the next two subsections we derive the gradients for stable neural flows via adjoint sensitivity analysis and the Lagrange multipliers method. We consider two types of cost functions: the \textit{terminal cost}, in which the cost only weights the terminal error, computed for $s=S$ and the \textit{back--propagated cost}, where, instead the cost is distributed on the whole domain $\Sa$. 
\subsection{Terminal Cost Gradients}
In this setting, $\ell_t := g(x(S),y_t)$ where $g$ is a smooth scalar function. The following hold:
\begin{proposition}[Terminal cost gradient]
    Consider a \textit{terminal} cost $\ell_t := g(x(S),y_t)$. Then, 
    \[
        \frac{d\ell_t}{dw} = \mu(0)
    \]
    where $\mu(s)\in\R^{n_w}$ satisfies the initial value problem  
    \[
        \begin{aligned}
            \dot\mu^\top(s) &= -\lambda^\top(s)\frac{\partial f}{\partial w}, &&\mu(S) = \mathbb{0}_{n_w}\\
            \dot\lambda^\top(s) &= -\lambda^\top(s)\frac{\partial f}{\partial x}, &&\lambda^\top(S) = \frac{\partial\ell_t}{\partial x(S)}
        \end{aligned}
    \]
    solved backward in $[0,S]$.
\end{proposition}
\proof
    Let us define a \textit{Lagrange multiplier} $\lambda(s)\in\R^{n_x}$ and let $\Lss_t$ be a perturbed loss function defined as
    \[
        \Lss_t := \ell_t(x(S)) - \int_0^{S} \lambda^\top(\tau)\left[\dot x(\tau) - f(u_t,x(\tau), w)\right]d\tau
    \]
    Since $\dot x - f(x,w)=0$ by construction, the integral term in $\Lss_t$ is always null and, thus, $\lambda(s)$ can be freely assigned while $\frac{d}{dw}\Lss_t = \frac{d}{dw}\ell_t$. For the sake of compactness we do not explicitly write the dependence on variables of the considered functions unless strictly necessary. Note that,
    \begin{equation}
        \begin{aligned}
            \int_0^{S} \lambda^\top&\dot xd\tau 
            = \lambda^\top(\tau)x(\tau)\big|_{0}^S - \int_0^{S} \dot\lambda^\top x d\tau
        \end{aligned}
    \end{equation}
    Hence,
    \begin{equation}
        \Lss_t = \ell_t(x(S)) - \lambda^\top(\tau)x(\tau)\big|_{0}^S + \int_0^{S} \left(\dot\lambda^\top x + \lambda^\top f\right)d\tau
    \end{equation}
    We can compute the gradient of $\ell_t$ with respect to $w$ as 
    \begin{equation*}
        \begin{aligned}
            \frac{d\ell_t}{dw}& = \frac{d\Lss_t}{dw} = \frac{\partial\ell_t}{\partial x(S)}\frac{d x(S)}{d w}\\
            & - \left(\lambda^\top(S)\frac{dx(S)}{dw}- \lambda^\top(0)\cancel{\frac{dx(0)}{dw}}\right) \\
            & + \int_0^S \left[\dot\lambda^\top\frac{d x}{dw} +  \lambda^\top\left(\frac{\partial f}{\partial w} + \frac{\partial f}{\partial x}\frac{dx}{dw} + \frac{\partial f}{\partial u_t}\cancel{\frac{du_t}{dw}}\right)\right]d\tau
        \end{aligned}
    \end{equation*}
    which, by reorganizing the terms, yields to 
    \begin{equation}\label{eq:fc_grad}
        \begin{aligned}
            \frac{d\ell_t}{dw} & = \left[\frac{\partial\ell_t}{\partial x(S)} - \lambda^\top(S)\right]\frac{dx(S)}{dw} + \\
            & + \int_0^S \left(\dot\lambda^\top + \lambda^\top\frac{\partial f}{\partial x}\right)\frac{d x}{dw}d\tau \\
            & + \int_0^S \lambda^\top\frac{\partial f}{\partial w}d\tau
        \end{aligned}
    \end{equation}
    Now, if $\lambda(s)$ satisfies the initial value problem
    \begin{equation}\label{eq:lm_ode}
        \begin{aligned}
            \dot\lambda^\top(s) = -\lambda^\top(s)\frac{\partial f}{\partial x},\quad\lambda^\top(S) = \frac{\partial\ell_t}{\partial x(S)}
        \end{aligned}
    \end{equation}
    to be solved backward in $[0,S]$; (\ref{eq:fc_grad}) reduces to 
    \begin{equation}\label{eq:fc_grad_rd}
        \begin{aligned}
            \frac{d\ell_t}{dw} &= \int_0^S\lambda^\top\frac{\partial f}{\partial w}d\tau\\
            &= - \int_S^0\left(\frac{\partial\ell_t}{\partial x(S)} - \int_S^\tau\lambda^\top(\zeta)\frac{\partial f}{\partial x}d\zeta\right)\frac{\partial f}{\partial w}d\tau
        \end{aligned}
    \end{equation}
    Note that (\ref{eq:fc_grad_rd}) can be computed by solving backward the system of ODEs 
    \begin{equation}\label{eq:adj_ode}
        \begin{aligned}
            \dot\mu^\top &= -\lambda^\top\frac{\partial f}{\partial w}, &&\mu(S) = \mathbb{0}_{n_w}\\
            \dot\lambda^\top &= -\lambda^\top\frac{\partial f}{\partial x}, &&\lambda^\top(S) = \frac{\partial\ell_t}{\partial x(S)}
        \end{aligned}
    \end{equation}
    Then, $\frac{d\ell}{dw} = \mu(0)$, proving the result.
\endproof
For the stable neural ODE (\ref{eq:snode}), system (\ref{eq:adj_ode}) becomes
\begin{equation}\label{eq:adj_ode_stable}
    \begin{aligned}
        \dot\mu^\top =  \lambda^\top\frac{\partial }{\partial w}\frac{\partial\varepsilon}{\partial x},\quad
        \dot\lambda^\top=  \lambda^\top\frac{\partial^2\varepsilon}{\partial x^2}
    \end{aligned}
\end{equation}
while, for the second order model (\ref{eq:s2node}), it holds
\begin{equation}\label{eq:adj_ode_2ord}
    \begin{aligned}
        \dot\mu^\top =  \lambda^\top\begin{bmatrix}
            \mathbb{O}_{n_q\times n_w}\\ -\partial_w\partial_q\varepsilon
        \end{bmatrix},\quad
        \dot\lambda^\top =  \lambda^\top
        \begin{bmatrix}
            \mathbb{O}_{n_q} & \mathbb{I}_{n_q}\\
            -\partial^2_q\varepsilon & -\alpha\mathbb{I}_{n_q}
        \end{bmatrix}.
    \end{aligned}
\end{equation}
\subsection{Back--Propagated Cost Gradients}
We can relax the results of the previous section to integral cost functions of type
\[
    \ell_t = \int_0^S g(x(\tau), y_t)d\tau
\]
In this context, similarly to the terminal cost case, the cost gradient are obtained by the following.
\begin{proposition}[Back--propagated cost gradient]\label{prop:bp_cost} Let $\ell_t = \int_0^S g(x(\tau), y_t)d\tau$. It holds,
    \[
        \frac{d\ell_t}{dw} = \mu(0)
    \]
    where $\mu$ satisfies the initial value problem  
    \begin{equation*}
        \begin{aligned}
            \dot\mu^\top(s) &= -\lambda^\top(s)\frac{\partial f}{\partial w},
           &\mu(S) = \mathbb{0}_{n_w}\\
            \dot\lambda^\top(s) &= -\lambda^\top(s)\frac{\partial f}{\partial x}- \frac{\partial g}{\partial x},  &\lambda(S) = \mathbb{0}_{n_x}
        \end{aligned}
    \end{equation*}
    solved backward in $[0,S]$.
\end{proposition}
\proof
    Proceeding in parallel to the terminal cost case yields
    \begin{equation}\label{}
        \begin{aligned}
            \frac{d\ell_t}{dw}& = \int_0^S\frac{\partial g}{\partial x}\frac{dx}{dw}d\tau\\
            & - \lambda^\top(S)\frac{dx(S)}{dw} \\
            & + \int_0^S \left[\dot\lambda^\top\frac{d x}{dw} +  \lambda^\top\left(\frac{\partial f}{\partial w} + \frac{\partial f}{\partial x}\frac{dx}{dw}\right)\right]d\tau
        \end{aligned}
    \end{equation}
    which leads to 
    \begin{equation}\label{}
        \begin{aligned}
            \frac{d\ell_t}{dw}& = - \lambda^\top(S)\frac{dx(S)}{dw} \\
            & + \int_0^S \left(\dot\lambda^\top + \lambda^\top\frac{\partial f }{\partial x} +\frac{\partial g}{\partial x}\right)\frac{d x}{dw}d\tau \\
            & + \int_0^S \lambda^\top\frac{\partial f}{\partial w}d\tau
        \end{aligned}
    \end{equation}
    Therefore, if $\lambda(s)$ satisfies
    \begin{equation}
        \begin{aligned}
            \dot\lambda^\top = \lambda^\top\frac{\partial f }{\partial x} - \frac{\partial g}{\partial x},\quad
        \lambda(S) = \mathbb{0}_{n_x}
        \end{aligned}
    \end{equation}
    we obtain:
    \begin{equation}\label{eq:bp_grad_rd}
        \begin{aligned}
            \frac{d\ell_t}{dw} &= \int_0^S\lambda^\top\frac{\partial f}{\partial w}d\tau\\
            & = -\int_S^0\left[\int_S^\tau\left(\lambda^\top(\zeta)\frac{\partial f}{\partial x} + \frac{\partial g}{\partial x}\right)d\zeta\right]\frac{\partial f}{\partial w}d\tau
        \end{aligned}
    \end{equation}
    Again, $d\ell/dw$ can be recovered as $\mu(0)$ by solving backward the ODE 
    \begin{equation}\label{eq:adj_ode_bp}
        \begin{aligned}
        \dot\mu^\top(s) &= -\lambda^\top(s)\frac{\partial f}{\partial w},
           &\mu(S) = \mathbb{0}_{n_w}\\
            \dot\lambda^\top(s) &= -\lambda^\top(s)\frac{\partial f}{\partial x}- \frac{\partial g}{\partial x},  &\lambda(S) = \mathbb{0}_{n_x}
        \end{aligned}
    \end{equation}
\endproof
\subsection{Gradients with Respect to Integration Bound}
It is also possible to optimise for the integration bound\footnote{Which can be interpreted, from an optimal control point of view, as the horizon of the control problem.} $S$ with GD iterates as in (\ref{eq:gd}) alongside $w$. In the case of the terminal cost loss, we have
%
\begin{equation}
    \begin{aligned}
        \frac{d\ell_t}{dS} &= \frac{\partial\ell_t}{\partial x(S)}\frac{d x(S)}{dS}=\frac{\partial\ell_t}{\partial x(S)}\frac{d}{dS}\int_0^Sf(x,w)d\tau\\
        &= \frac{\partial\ell_t}{\partial x(S)}f(x(S),w),
    \end{aligned}
\end{equation}
where the Leibniz integral rule has been used in the last equality. Hence, for stable neural ODEs the cost gradient is directly correlated to the gradient of the energy stored in the system at the end of the integration, i.e.
\begin{equation}\label{eq:S_grad}
    \frac{d\ell_t}{dS} = -\frac{\partial \ell_t}{\partial x(S)}\partial_x\varepsilon(x(S),w)
\end{equation}
In the case of back--propagated cost, instead, we have
\begin{equation}
    \begin{aligned}
        \frac{\partial\ell_t}{\partial S} &= g(x(S)).
    \end{aligned}
\end{equation}
\subsection{Gradients with respect to input and output projections}
In case $h_u$ and $h_y$ depend on two sets of parameters $v_u$ and $v_y$, which we want to optimise with GD alongside $w$, we need to compute their respective gradients. In the terminal cost case, it holds that
\[
    \ell_t = g(x(s)) = g\left[h_y\left(h_u(u_t) + \int_0^Sf(x(\tau))f\tau\right)\right]
\]
and, therefore,  
\[
    \begin{aligned}
        \frac{d\ell_t}{dv_u} =\frac{\partial g}{\partial h_y}\frac{\partial h_y}{\partial h_u}\frac{\partial h_u}{\partial v_u},\quad        \frac{d\ell_t}{dv_y} =\frac{\partial g}{\partial h_y}\frac{\partial h_y}{\partial v_y}
    \end{aligned}
\]
Moreover, in the case of back--propagated cost, the gradients with respect to $v_u$ can be computed as
\[
    \begin{aligned}
        \frac{d\ell_t}{dv_u} &=\frac{d}{dv_u}\int_0^S g(x(\tau))d\tau\\
    &= \int_0^S\frac{d}{dv_u} g\left[h_u(u_t, v_u) + \int_0^\tau f(u_t,x(\zeta),w)d\zeta\right]d\tau\\
    &= 
    \int_0^S\frac{\partial g}{\partial  h_u}\frac{\partial h_u}{\partial v_u}d\tau = 
    \int_0^S\frac{\partial g}{\partial  h_u}d\tau\frac{\partial h_u}{\partial v_u}
    \end{aligned}
\]
thus resulting in $\frac{d\ell_t}{d v_u} = \chi(0)\frac{\partial h_u}{\partial v_u}$, where $\chi(s)$ satisfies
\[
    \dot \chi^\top(s) = -\frac{\partial g}{\partial h_u},~\chi^\top(S) = \mathbb{0}_{n_x}
\]
\subsection{On Training Regularisers}
While we enforce stability to stable neural flows by structuring the vector field as the negative gradient of a bounded energy functional, we have no guarantees that the system approaches steady--steady at the end of a bounded depth domain. Therefore, it would be beneficial to introduce suitable soft constraints as additive term to the cost $\ell_t$ to accelerate convergence to steady--state.
To this end, we can augment the loss with a terminal cost term proposed by \cite{massaroli2020dissecting}:
\begin{equation}\label{eq:reg_loss}
    \begin{aligned}
        \ell^{\star}_t &= \ell_t + \frac{\gamma}{2}\|f(u_t,x(S),w)\|_2^2  \\
        &= \ell_t + \frac{\gamma}{2}\left\|\partial_x\varepsilon(u_t,x(S), w)\right\|_2^2 
    \end{aligned}
\end{equation}
with $\gamma\in\R^+$. Note that this regularization term may be also successfully used in standard \textit{vanilla} models to encourage regularity and stability of the flows \cite{massaroli2020dissecting}. 
\section{EXPERIMENTAL EVALUATION}
In all the following experiments, the ODEs have been solved with a Dormand--Prince adaptive--step solver and absolute and relative tolerances set to $10^{-6}$ while the scalar $\gamma$ for the regularized loss (\ref{eq:reg_loss}) has been set to $10^{-2}$.  
\subsection{Function approximation}
We consider the approximation of the function $y = - u$ ($u\in\R$). While standard, neural ODEs \cite{chen2018neural} cannot tackle this simple problem \cite{dupont2019augmented,massaroli2020dissecting}, we show how a stable model dependent on $u_t$ is able to learn a proper family of energy functions whose steepest direction is followed. In this case, $\varepsilon$ was chosen as multi--layer perceptron taking $x,~u_t$ ad inputs with layers $2,16,16,1$ and \textit{hyperbolic tangent} (tanh) activation at each layer but the last one. The output is then squared to enforce lower boundedness. The training was performed by uniformly sampling $u_t \in \mathcal{U} := [-1,1]$ and computing $y_t = -u_t$. Figure~\ref{fig:surf_1d} shows how the state set $\mathcal{U}$ is adjusted by the model to lie on a lower energy configuration of the energy functional $\epsilon$. This minimum energy configuration is shaped by $f$ during optimization to correspond to the desired function $y = -u$. Figure~\ref{fig:crossing_1d} offers a different perspective on the same task and highlights the ability of $f$ to learn crossing flows.
\begin{figure*}[t]
    \centering
    \begin{tikzpicture}[scale=0.9, every node/.style={scale=0.9}]
\begin{groupplot}[group style={group size=3 by 1}]

\nextgroupplot[
	title = {$s=0$},
	width = .3\linewidth,
	height = 3cm,
	colormap  = {whiteblack}{color(0cm)=(white);color(1cm) = (black)},
    view/h=40,
    xlabel=$x$, ylabel=$u_t$, 
    zlabel=$\varepsilon \left(x\mathpunct{,} u_t\right)$,
]
\addplot3[surf,mesh/ordering=y varies,
    shader=interp]
    table {energy_1D.txt};
\addplot3[ultra thick, smooth, mark=none, color=orange] table {tr_0.txt};
\nextgroupplot[
	title = {$s=0.5$},
	width = .3\linewidth,
	height = 3cm,
	colormap  = {whiteblack}{color(0cm)=(white);color(1cm) = (black)},
	view/h=40,
	xlabel=$x$, ylabel=$u_t$, 
	zlabel=$\varepsilon \left(x\mathpunct{,} u_t\right)$,
]
\addplot3[surf,mesh/ordering=y varies,
shader=interp]
table {energy_1D.txt};
\addplot3[ultra thick, smooth, mark=none, color=orange] table {tr_1.txt};
\nextgroupplot[
	title = {$s=1$},
	width = .3\linewidth,
	height = 3cm,
	colormap  = {whiteblack}{color(0cm)=(white);color(1cm) = (black)},
	view/h=40,
	xlabel=$x$, ylabel=$u_t$, 
	zlabel=$\varepsilon \left(x\mathpunct{,} u_t\right)$,
]
\addplot3[surf,mesh/ordering=y varies,
shader=interp]
table {energy_1D.txt};
\addplot3[ultra thick, smooth, mark=none, color=orange] table {tr_2.txt};

\end{groupplot}
\end{tikzpicture}
    \vspace{-5mm}
    \caption{Snapshots of the state evolution through the depth domain over the learned energy functional.}
    \label{fig:surf_1d}
\end{figure*}
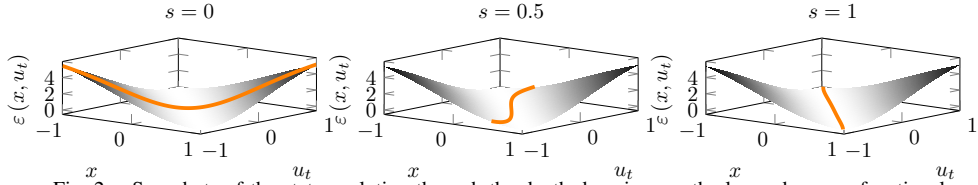
\begin{figure}
    \centering
    \input{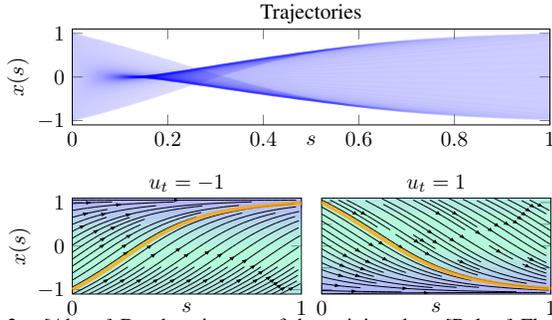}
    \vspace{-5mm}
    \caption{[Above] Depth trajectory of the training data. [Below] Flows over the data--dependent learned vector field for $u_t=-1$ and $u_t=1$}
    \label{fig:crossing_1d}
\end{figure}
\subsection{Nonlinear classification via stable neural flows}
    We tested several models with different two-dimensional nonlinear classification tasks, including the \textit{half--moons} and \textit{three spirals} datasets. The objective of each neural ODE model tested was then steering the input points of different classes towards linearly separable manifolds, so that the linear output projection map could minimize the chosen cost function. For each problem, we sampled the classes in the input space and added Gaussian noise. The \textit{output projection} map $h_y:\R^{n_x}\rightarrow\R^{n_y}$ was chosen as a linear affine map $\hat y_t = h_y(x):= W_y^\top x + b_y$ where $v_y:=\text{vec}({W_y,b_y})\in\R^{(n_x+1)n_y}$ has been optimized together with the models' parameters $w$. In both cases $\vep$ was parametrised by a multilayer perceptron with two hidden layer of 32 units each and tanh activation. The scalar output was then filtered by a sigmoid function to enforce lower boundedness of $\vep$. 
    
\begin{figure}[b]
    \centering
    \includegraphics{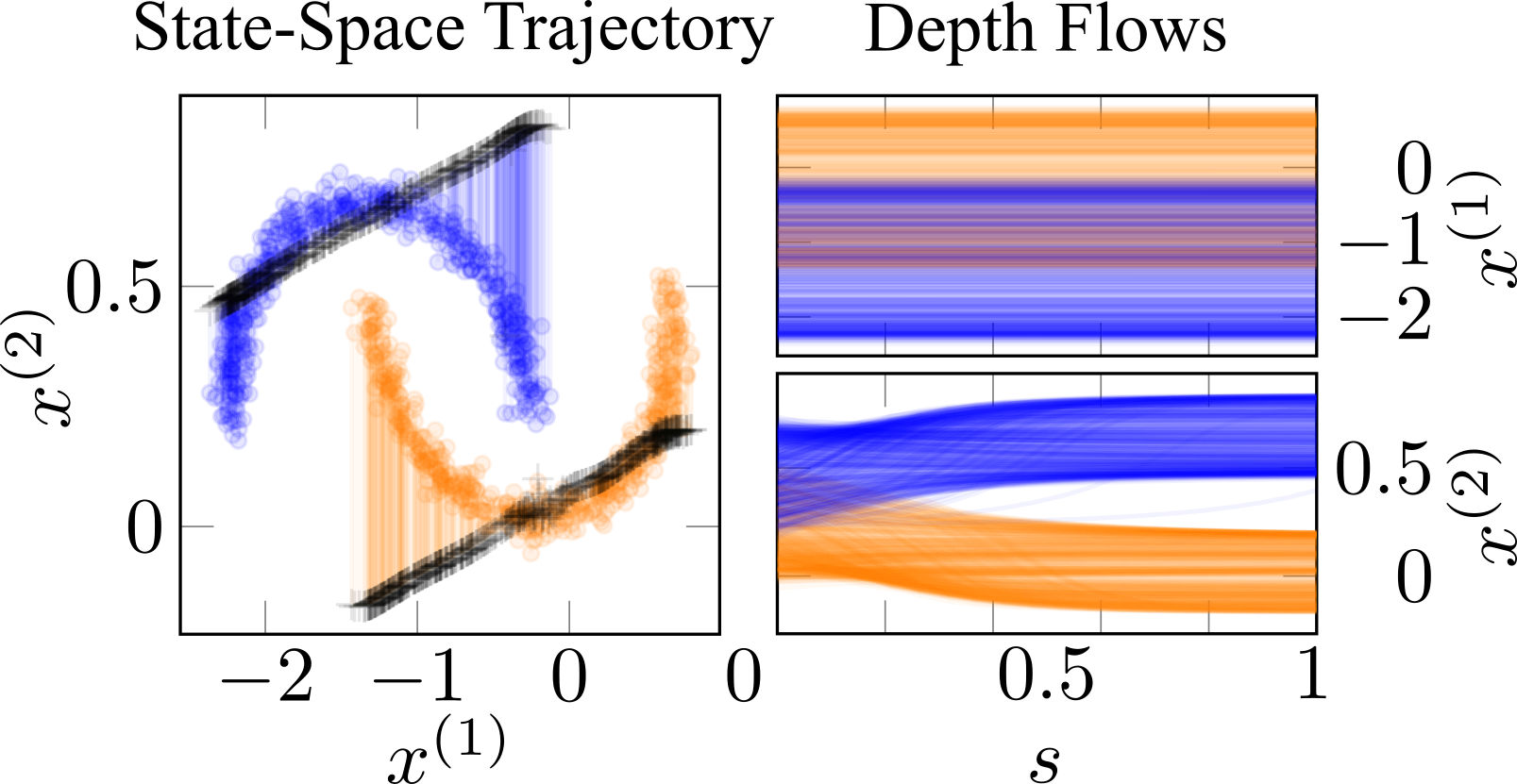}
    \vspace{-4mm}
    \caption{\textit{Half--Moons} classification task carried out with stable neural flows where $\vep$ is $u_t$--independent.}
    \label{fig:half_moon_traj}
\end{figure}

\begin{figure}[b]
    \centering
    \includegraphics{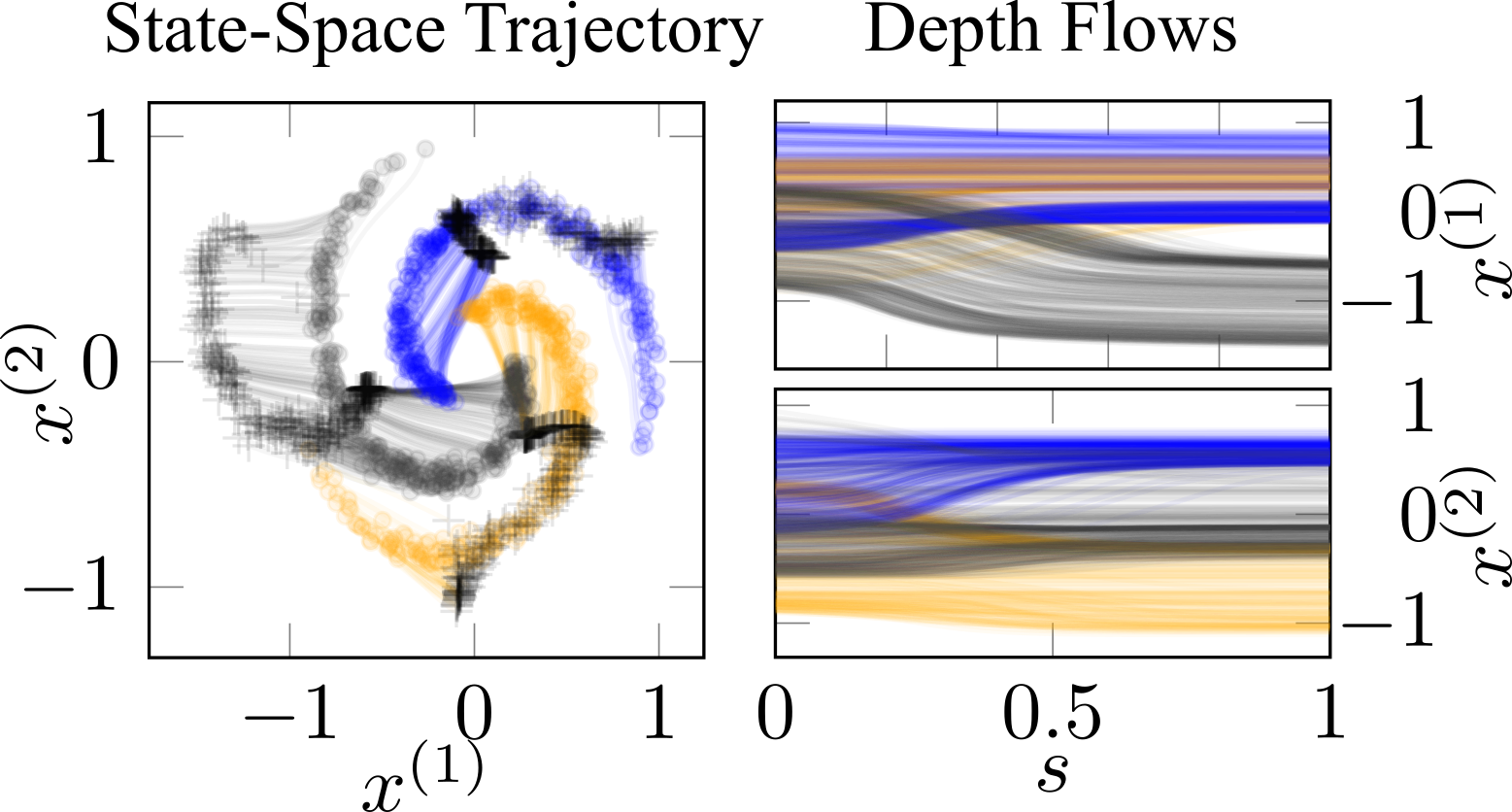}
    \vspace{-4mm}
    \caption{\textit{Three--spirals} classification carried out with stable neural flows and \textit{data-dependent} energy function $\vep$.}
    \label{fig:spirals_traj}
\end{figure}

    We evaluate different variants of the proposed method. For half--moons, we employ a stable neural flow in port--Hamiltonian form (\ref{eq:ph}) with $A := -\diag(|a_1|, |a_2|)$, $w_A:=(a_1, a_2)$ but with no explicit dependence of $\vep$ from $u_t$ (data--independent). The model is equipped with an $\R^2\rightarrow\R^2$ \textit{input projection} map $h_u(x):= W_u^\top u_t + b_u$, with parameters, $v_u:=\text{vec}(W_u, b_u)\in\R^{6}$ included in the optimization procedure. We compute the quadratic cost of outputs $\hat y_t$ and corresponding labels $y_t \in \{0,1\}$. The three spirals experiments, on the other hand, involve the use of a data--dependent stable neural flow (\ref{eq:snode}) optimized to minimize the \textit{cross-entropy} \cite{shore1982minimum} cost of $\hat y_t$ and \textit{one--hot} encoded labels $y_t \in \R^{3}$.
    Figures~\ref{fig:half_moon_traj} and ~\ref{fig:spirals_traj} show that the model correctly steers input--data towards such linearly separable clusters, indicated in black. The depth--flows converge to the desired values before $S := 1$, confirming stability.   
\section{CONCLUSION AND FUTURE WORK}
We enhance \textit{neural ordinary differential equations} with intrinsic stability properties. Key to \textit{stable neural ODEs} is
replacing an unconstrained vector field with an energy manifold whose gradient drives the data--flows. Furthermore, the proposed model class is augmented with \textit{ad hoc} regularizers designed to accelerate convergence to steady--state. Distilling an energy representation of the input manifold provides provides a robust approach to continuous--depth deep learning and naturally paves the way to future work involving observers and controllers based on neural ODEs.
\bibliographystyle{unsrt}
\bibliography{biblio.bib}
\end{document}